\documentclass[11pt]{article}

\usepackage{amsmath,amssymb,amsfonts,latexsym,lscape,rawfonts}
 \usepackage{caption} 
\usepackage{color}
\usepackage{algorithm}
\usepackage{algorithmic} 
\usepackage{booktabs} 
\usepackage{multirow} 
\usepackage{epstopdf}
\usepackage{graphicx}
\usepackage{url}

\DeclareCaptionLabelFormat{algnonumber}{Subroutine} 

\newcommand{\blue}[1]{{#1}}
\newcommand{\red}[1]{{#1}}

\begin{document}


\title{Efficient online learning for large-scale peptide identification }
\author
{ Xijun Liang\,$^{\text{1}}$,
  Zhonghang Xia\,$^{\text{2}}$,
  Yongxiang Wang\,$^{\text{1}}$,\\
  Ling Jian\,$^{\text{1,}*}$,
  Xinnan Niu\,$^{\text{3}}$
 and Andrew Link\,$^{\text{3}}$
 }
 \maketitle
 
\noindent
\textit{$^{\text{\sf 1}}$ College of Science, China University of Petroleum,
Qingdao, China, 266580 \\
$^{\text{\sf 2}}$ Department of Computer Science, Western Kentucky University,
Bowling Green, KY 42101
 and \\
$^{\text{\sf 3}}$ Dept. of Pathology, Microbiology and Immunology,
Vanderbilt University School of Medicine, Nashville, TN 37232 \\
}



{\Large \textbf{Abstract}}
\medskip

{\textbf{Motivation:} Post-database searching is a key procedure in peptide identification with tandem mass spectrometry (MS/MS) strategies for refining \red{peptide-spectrum} matches (PSMs) generated by database search engines.
Although many statistical and machine learning-based methods have been developed
to improve the accuracy of peptide identification, the challenge remains on large-scale datasets and
 datasets with \red{an} extremely large proportion of false positives (hard datasets).
A more efficient learning strategy is required for improving the performance of peptide identification on challenging datasets.

\textbf{Results:} In this work,  we present an online learning method
 to  conquer the   challenges remained for exiting peptide identification algorithms.
We propose a cost-sensitive learning model by using different loss functions for decoy and target PSMs respectively.
A larger penalty \red{for} wrongly selecting decoy PSMs than that \red{for} target PSMs,
and thus the new model can reduce its false discovery rate on hard datasets.
Also, we design an online learning algorithm, OLCS-Ranker,
to solve the proposed learning model.
Rather than taking all training data samples all at once, OLCS-Ranker iteratively \red{feeds} in only one training sample into the learning model at each round. As a result, the memory requirement is significantly reduced for large-scale problems.
Experimental studies show that
OLCS-Ranker outperforms  benchmark methods,
such as CRanker \blue{and Batch-CS-Ranker},  in terms of accuracy and stability.
Furthermore, OLCS-Ranker is 15--85   times faster  than  CRanker method on large datasets.

\textbf{Availability and implementation:}  OLCS-Ranker software is available at no charge for non-commercial use at
 https://github.com/Isaac-QiXing/CRanker.
 
\textbf{Contact:}  { liangxijunsd@163.com or  bebetter@upc.edu.cn}

}

\section{Introduction}

Tandem mass spectrometry (MS/MS)-based strategies are presently the method of choice for large-scale protein identification due to its high-throughput analysis of biological samples. With database sequence searching method,
 a huge number of peptide spectra generated from MS/MS experiments are routinely searched by using a search engine, such as SEQUEST, MASCOT or X!TANDEM, against theoretical fragmentation spectra  derived from target databases or  experimentally observed spectra
  for \red{peptide-spectrum} match (PSM).


    A number of computational methods and error rate estimation procedures
    \cite{Nesvizhskii2010A}
    have been proposed to improve the accuracy of target PSMs.
 In the early trials, empirical filters \cite{Link1999}
 were developed to choose the target PSMs, in which those above the specified
thresholds are accepted as correct and those below the
thresholds are assumed to be incorrect.
With the absence of robust
statistical and computational methods,  these filtering methods could not achieve  satisfactory  identification results, especially  on the datasets
containing significant numbers of false positives.

Advanced statistical and machine learning approaches have been studied for improving the accuracy of discrimination of correct and incorrect PSMs.
 Among those machine learning-based tools, PeptideProphet \cite{Andrew2002Empirical} and Percolator \cite{K2007Semi} are two popular ones using semi-supervised learning.
    PeptideProphet employs the expectation maximization method to compute the probabilities of correct and incorrect PSM, based on
    the assumption that the PSM data are drawn from a mixture of two Gaussian distributions which generate samples of the correct and incorrect PSMs.
The learning model of PeptideProphet was extended in \cite{Choi2007}
by  incorporating decoy PSMs into a mixture
probabilistic model
 at the estimation step of the expectation maximization.
  Percolator starts the learning process with a small set of trusted correct PSMs and  incorrect PSMs selecting from  a decoy database,
    and it iteratively adjusts its learning model to fit the dataset and ranks the PSMs according to confidence on the PSM.

 Another category of machine learning-based methods \red{uses} supervised learning and \red{formulates} peptide identification
as an optimization problem.
  In \cite{Spivak2009}, 
  a fully supervised   method
   is proposed to improve the performance of Percolator.
    Two types of discriminant functions, linear functions and two-layer neural networks, are compared.
    The two-layer  neural networks is a nonlinear discriminant function which adds
    lots parameters of hidden units. As expected, it achieves better identification performance than
    the model with linear discriminant function \cite{Spivak2009}.
    CRanker \cite{Liang2015cranker} is a method
     that  employs
      kernel-based SVM to formulate the peptide identification
    problem as  an optimization problem. 
     Although    CRanker has 
       shown efficiency   compared with benchmark approaches, PeptideProphet and Percolator,
     it  
     could not efficiently deal with the large-scale PSM datasets because of the storage
      of large kernel matrix and computation complexity.

Although these  advanced   post-database searching approaches have dramatically improved the accuracy of peptide identification,
two big challenges remain in practical implementation:

\begin{enumerate}

\item the performance of the algorithms degrades on the PSM datasets having \red{an} extremely large proportion of  false positives (called \textit{``hard dataset''});

\item \red{a} huge amount of computational time and resources are required for large-scale datasets, resulting in
\red{a} heavy burden on computation.
\end{enumerate}
A number of works have attempted to conquer the two challenges.
 Authors in \cite{Wang2014Integrating} integrate auxiliary information to improve the identification performance for the ``hard datasets'' challenge.
  Moreover,  MSFragger \cite{Kong2017MSFragger}
 empowers the open database search concept and includes all the modified forms of the peptides to
 improve the matching quality.
For the other challenge, cloud computing platform is used in \cite{Slagel2015Cloud} to tackle the intense
memory requirement.


 In this work, we aim at tackling the two challenges by using efficient 
  optimization techniques.
For the challenge of ``hard dataset'',
we first \red{extend}
   CRanker \cite{Liang2015cranker} model to a cost-sensitive CRanker (CS-Ranker) by using different loss functions for decoy and target PSMs respectively. The CS-Ranker model gives a larger penalty \red{for} wrongly selecting decoy PSMs than that \red{for} target PSMs,
        which reduces the model's false discovery rate while increases its true positive rate.
        Second, we designed an online algorithm randomly selecting the  PSM samples and add them into \red{the} training process.
       As a result, the training model is less prone to converging to \red{poor local minima}, avoiding extremely bad identification results.

 The challenge of \red{the} requirement of computational resource comes from the large-sized PSM datasets. CRanker and other kernel-based  batch learning algorithms need to load the entire kernel matrix into memory, and thus the memory requirement can be very intense during the training process.
Also, parameter selection for CRanker and most machine learning-based models is very time-consuming.
For choosing a set of appropriate parameters, it usually takes CRanker
 dozens of hours  to determine the discriminant function  on large-sized datasets.

 In order to reduce the high demand on computer memory, we construct a new classification model, CS-Ranker, by incorporating the  C-SVM model into  an online algorithm for CS-Ranker (OLCS-Ranker) which trains PSM data samples one by one and  uses an active set to
 keep only those PSMs
 effective to the discriminant function. In this way, memory requirement and total training time can be dramatically reduced.


Experimental studies  have shown that
 OLCS-Ranker
 outperforms CRanker on the tested datasets  in terms of accuracy and stability while
 it reports a list of target PSMs comparable to PeptideProphet and Percolator.
More importantly, OLCS-Ranker is $15\sim 85$  times faster on large datasets
 than CRanker.




\section{Methods}


\subsection{Cost-Sensitive Ranker model}


In this section, we present a cost-sensitive classification  model which
 extends   CRanker \cite{Liang2015cranker} model by using different loss functions for decoy and target PSMs respectively.

Identification of correct target PSM can be formulated as a classification problem.
Let $\Omega=\{x_{i}^{},y_{i}^{}\}_{i=1}^{n} \subseteq R_{}^{q} \times \{-1,1\}$ be a set of $n$ PSMs,
where $x_{i}^{} \in R_{}^{q}$ represents its $i$-th PSM record with  $q$ attributes, and
$y_{i}^{} = 1$ or  $-1$ is the corresponding label indicating a target  or  decoy PSM.
The identification task is to train a discriminant function for
filtering out the correct PSMs from the target PSMs (ones with labels ``$+1$'').

Let
$
  \Omega_{+}^{} = \{j \,|\, y_{j}^{} = 1 \}, \quad
  \Omega_{-}^{} = \{j \,|\, y_{j}^{} = -1 \}.
$
A commonly used Support Vector Machine (SVM) \cite{Burges1998},  C-SVM, assigns the
 PSM labels according to a discriminant function $f$ given by the
 following optimization model:
\begin{equation}\label{EQ_CSVM_abstract}
  \min_{w,b} \  \frac{1}{2}\|w\|_{}^{2} + C\sum_i h(y_{i}f(x_i))
\end{equation}
where $C>0$ is the weight of the experiential loss,
  $h(t) = \max(0,1-t )$ is the hinge loss function, and
$f(x_i) = \langle w, \phi(x_i)\rangle +b$ is the value of discriminant function at $x_i$.
$\phi(\cdot)$ is a feature mapping,
 Here, we set offset $b \equiv 0$.

While class labels in a standard classification problem are trustworthy,  a large number of ``$+1$''
labels in PSM identification   are not correct.
 CRanker\cite{Liang2015cranker} introduces  weight $\theta_i \in [0,1]$,
 for each PSM sample $(x_i,y_i)$ to indicate the degree of the reliability of the label $y_i$.
Particularly, $\theta_i = 1$ indicates that label $y_i$ is definitely correct,
$\theta_i = 0$ indicates that it is definitely incorrect,
and $\theta_i \in (0,1)$ indicates that label $y_i$ is probably   correct.
In fact, all ``$-1$'' labels (decoy PSMs) are correct, and thus $\theta_i = 1$ for all $i\in \Omega_{-}^{}$.
 Following the idea of CRanker we propose to solve  the following optimization problem

\begin{equation}\label{EQ_cranker_2var}
  \begin{array}{cl}
     \min_{w,\theta}^{} & \frac{1}{2}\|w\|^2 + C \sum_{i=1}^{n} \theta_{i}^{} h(y_{i}^{}f(x_i) ) -
      \lambda \sum_{i=1}^{n} \theta_{i}^{}   \\
    \textnormal{s.\,t.}
                & \theta_{i}^{} = 1, \ i \in \Omega_{-}^{}, \\
                &  0\leq \theta_{i}^{} \leq 1,\  i \in \Omega_{+}^{}, \\
  \end{array}
\end{equation}
where $\lambda>0$ is the weight to encourage the model to identify more correct PSMs.
 As shown in  \cite{Liang2017SPL, Meng2015objective},
a larger value of parameter $\lambda$
selects more PSMs into training process.



Note that if the discriminant function assigns ``$+1$'' label to a decoy PSM, then we know for sure that the label assignment is wrong.
 In this case, the learning error is more likely caused by the model itself rather than the quality of data sample, and hence we should give the loss function  a large penalty.
 On the other hand, if  a target is classified as negative and assigned  label ``$-1$'',
    we are not even sure whether the label assignment is correct, and thus we consider  a small penalty for the loss function.
    Based on these observations, we incorporate the new penalty policy into
    model (\ref{EQ_cranker_2var}) and the new model is described as follows:

\begin{equation}\label{EQ_cranker_unequal_cost}
  \begin{array}{cl}
     \min_{w,\theta}^{} & \frac{1}{2}\|w\|^2 + C_1 \sum_{i\in\Omega_{-}^{}}^{ } \theta_{i}^{} h(y_{i}^{}f(x_i) )\\
       & \quad + C_2 \sum_{i\in\Omega_{+}^{}}^{ } \theta_{i}^{} h(y_{i}^{}f(x_i) )
      -  \lambda \sum_{i=1}^{n} \theta_{i}^{}   \\
    \textnormal{s.\,t.}
                & \theta_{i}^{} = 1, \ i \in \Omega_{-}^{}, \\
                &  0\leq \theta_{i}^{} \leq 1,\  i \in \Omega_{+}^{}, \\
  \end{array}
\end{equation}
where $C_1>0$, $C_2>0$ are  cost weights for the losses of the   decoys and  targets, respectively.
Model (\ref{EQ_cranker_unequal_cost}) is named \textbf{cost-sensitive ranker} model and denoted by \textbf{CS-Ranker}.
As we choose a larger penalty for decoy losses,  $C_1^{} \geq C_2^{}$ always holds.

\subsection{The batch convex-concave procedure for solving CS-Ranker }

%

 In this section,  we present a batch  algorithm to solve  the CS-Ranker model
     by leveraging the DC (\textbf{d}ifference of two \textbf{c}onvex functions) structure of  (\ref{EQ_cranker_unequal_cost}).



It can be shown by \cite{Meng2015objective} that
 if a pair of $w^*\in R^n$ and $\theta^{*} \in R^n$ is an optimal solution
to   CS-Ranker model (\ref{EQ_cranker_unequal_cost}), then
$w^*$ is also an optimal solution of the following model,
\begin{equation}\label{EQ_cranker_1var}
     \min_{w}^{} \, \frac{1}{2}\|w\|^2
        + C_1 \sum_{i\in\Omega_{-}^{}}^{}  h(y_{i}^{}f(x_i) )
        + C_2 \sum_{i\in\Omega_{+}^{}}^{} R_{s}^{}(y_{i}^{}f(x_i) )
\end{equation}
with  $R_{s}(t)  = \min(1-s,\max(0,1-t) )$,
$s = 1- \frac{\lambda}{C_{2}^{}}$, and vice versa.


    Since $
     R_s(t) = H_1(t) - H_s(t),
    $
     with      $H_s(t) = \max(0,s-t )$,
then model  (\ref{EQ_cranker_1var}) can be recast as
programming:
\begin{equation}\label{EQ_cranker_DC}
     \min \quad J(w) = J_{\textnormal{vex}}^{}(w) + J_{\textnormal{cav}}^{}(w),
\end{equation}
where
\begin{equation}\label{EQ_J_cav_J_vex}
\begin{array}{l}
J_{\textnormal{vex}}^{}(w)= \frac{1}{2}\|w\|^2
         + C_{1}^{} \sum_{i\in\Omega_{-}^{}}^{ }  H_1(y_{i}^{}f(x_i) ) \\
   \qquad     + C_{2}^{} \sum_{i\in\Omega_{+}^{}}^{ }  H_1(y_{i}^{}f(x_i) ), \\
J_{\textnormal{cav}}^{}(w) = - C_{2}^{} \sum_{i\in\Omega_{+}^{}}^{} H_{s}^{}(y_{i}^{}f(x_i) ).
\end{array}
\end{equation}
 $J_{\textnormal{vex}}^{}(\cdot)$ and  $J_{\textnormal{cav}}^{}(\cdot)$
 are convex and concave functions respectively.
Model\,(\ref{EQ_cranker_DC}) can be
solved by
 a standard  Concave-Convex Procedure (CCCP) \cite{yuille2003},
    which iteratively solves subproblems
     \begin{equation}\label{EQ_ite_CCCP}
        w_{}^{k+1} = \textnormal{argmin}_{w}^{} \quad J_{\textnormal{vex}}^{}(w) + J_{\textnormal{cav}}^{\prime}(w^k)\cdot w
     \end{equation}
     with initial $w^0$.
The Lagrange dual  of (\ref{EQ_ite_CCCP}) can be deduced as follows:
\begin{equation}\label{EQ_ite_CCCP_dual}
    \begin{array}{cl}
    \max_{\alpha}^{} &  G(\alpha) = -\frac{1}{2}\sum_{i,j}^{}\alpha_{i}^{} \alpha_{j}^{} k(x_i,x_j)
                + \langle \alpha,y \rangle +   \sum_{i\in \Omega_{+}}^{} C_2 \eta_{i}^{k}   \\   
    \textnormal{s.\,t.} &   A_i \leq \alpha_i \leq B_i, \quad, i=1,\ldots, n \\
            & A_i = \min(0,C_{1}^{} y_i),\ i\in \Omega_{-}^{} \\
            & B_i = \max(0,C_{1}^{} y_i),\ i\in \Omega_{-}^{} \\
            & A_i = \min(0,C_{2}^{}y_i)-C_{2}^{}\eta_{i}^{}y_i,\ i\in \Omega_{+}^{} \\
            & B_i = \max(0,C_{2}^{}y_i)-C_{2}^{}\eta_{i}^{}y_i,\ i\in \Omega_{+}^{} \\
    \end{array}
\end{equation}
where $\eta_{i} =
\left\{
\begin{array}{cl}
1, & \textnormal{ if } y_i f^{}(x_i) <s,\\
0, & \textnormal{ otherwise }.
\end{array}
\right.$

  Based on  the CCCP framework,
   we   solve CS-Ranker with Algorithm\,\ref{Alg_Batch_CS_Ranker} as follows.

\begin{algorithm}[]
            \caption{Batch  algorithm for solving CS-Ranker (Batch-CS-Ranker)}\label{Alg_Batch_CS_Ranker}
\algsetup{
    linenosize=\small,
    linenodelimiter=.
}

\begin{algorithmic}[1]
\STATE             Initialize $\eta_{i}^{} =0$, $i \in \Omega_{+}^{}$, $k=0$.
\REPEAT
\STATE Solve the quadratic programming (\ref{EQ_ite_CCCP_dual}), and the optimal solution is set as
    $\alpha_{}^{k+1}$.

\STATE update $\eta_{}^{k+1}$:
  $ \eta_{i}^{k+1} =
\left\{
\begin{array}{cl}
1, & \textnormal{ if } y_i f^{k+1}(x_i) <s,\\
0, & \textnormal{ otherwise },
\end{array}
\right.
  $
    $i\in \Omega_{+}^{}$, $f_{}^{k+1}(x_i) = \sum_{j=1}^{n} \alpha_{j}^{k+1} k(x_j,x_i)$.

\STATE Update $A_i$ and $B_i$:  
  $ A_i = \min(0,C_{2}^{}y_i)-C_{2}^{}\eta_{i}^{k}y_i$,
  $ B_i = \max(0,C_{2}^{}y_i)-C_{2}^{}\eta_{i}^{k}y_i$, $i\in \Omega_{+}^{}$.

\STATE set $k\longleftarrow k+1$.
 \UNTIL{convergence of $\alpha_{k}$}

\end{algorithmic}
\end{algorithm}

 Algorithm\,2   iteratively executes    two main steps: 1) Solve the SVM quadratic programming (\ref{EQ_ite_CCCP_dual}) (Line 3). 2) Updates the bounds $A_i$ and $B_i$ with $i\in \Omega_{+}^{}$ (Line\,4--Line\,5).
 As the training set is fed into the algorithm once at the beginning, we name Algorithm\,2 \textbf{Batch-CS-Ranker}.




\subsection{The online learning algorithm for CS-Ranker model}\label{sec_online_cranker}

 Inspired by the work in \cite{Bordes2005,Ertekin2011}, we present  an \textbf{online CS-Ranker algorithm} for overcoming the two challenges remaining in the Batch-CS-Ranker algorithm, and name it \textbf{OLCS-Ranker}.
It is actually
an online version of the Batch-CS-Ranker algorithm, and
both algorithms solve CS-Ranker model\,(\ref{EQ_cranker_1var})
to train the discriminant function.
Different with the Batch-CS-Ranker algorithm taking the PSM samples all at once,
 OLCS-Ranker train the discrimination function in \red{an} iterative manner and
  adds only one PSM sample into the training process at each iteration. The PSM sample is randomly selected to help
  the solution of discrimination function not trap at a   local minimum, and the effectiveness has been observed in approaches such as stochastic gradient descent\cite{Bottou1991SGD}.
  Moreover, OLCS-Ranker maintains an active set to only keep indices of PSMs that determine the discriminant function
in model training , and the PSMs that do not affect the discriminant function are discarded.
As a result, the cost of memory and computation is minimized.

\subsubsection{Online Algorithm for Solving CS-Ranker}


The implementation of OLCS-Ranker is depicted in Algorithm\,\ref{alg_OLCS_Ranker}.
 Particularly,
given  a chosen PSM sample (Line\,3),
OLCS-Ranker updates bounds $A_j$, $B_j$, for all $j \in \Omega_{+}^{}\cap S$ (Line\,4 -- Line\,7),
 and call subroutines PROCESS and REPROCESS to
solve dual programming (\ref{EQ_ite_CCCP_dual})
 with training samples in active set $S$ (Line\,8--Line\,12).
Periodically, the algorithm call subroutine CLEAN    to remove part of
 redundant instances from the kernel expansion (Line 13).
 The iteration terminates when all the training PSMs has been chosen for training.

\begin{algorithm}

\caption{ Online CS-Ranker algorithm for solving model\,(\ref{EQ_cranker_1var})}
\label{alg_OLCS_Ranker}
\begin{algorithmic}[1]
 \REQUIRE PSM samples $\{x_i,y_i\}_{i=1}^n$.
 \ENSURE solution $\alpha\in R^n$.

\noindent \textbf{Parameters:}

\noindent $M$: minimum number of PSMs in the active set $S$;

\noindent $\tau>0$: the tolerance to solve the dual programming (\ref{EQ_ite_CCCP_dual});

\noindent $\mu_{\textnormal{safe}}^{},\ \mu_{\textnormal{safe-target}}^{}  $:  thresholds to select candidate PSMs.

\STATE Set $\eta \leftarrow 0$, $ \alpha \leftarrow 0$, $S\leftarrow \emptyset$.

\FOR{$i_0 \in \{1,2,\ldots, n\}$}

\STATE Randomly select a training PSM sample $(x_{i_0}^{},y_{i_0}^{})$.

 \STATE  Update bounds $A_j$, $B_j$, $\forall j \in \Omega_{+}\cap S$:

\STATE $S\leftarrow S\cap \{i_0\} $;

\STATE Set $ \eta_{j}^{} =
\left\{
\begin{array}{cl}
1, & \textnormal{ if } y_j \hat{f}^{}(x_j) <s \textnormal{ and } |S|>M,\\
0, & \textnormal{ otherwise },
\end{array}
\right.
  $,
   $j\in \Omega_{+}^{}\cap S$, $\hat{f}(x_j) = \sum_{s\in S}^{}\alpha_s k(x_s,x_j)$;

\STATE Update bounds
$
        A_j = \min(0,C_2 y_j) - C_2 \eta_{j}^{k}y_j,\  \
        B_j = \max(0,C_2 y_j) - C_2 \eta_{j}^{k}y_j,\ j\in \Omega_{+}^{}\cap S.
$

\STATE  Call PROCESS().

\STATE
  exitFlag $\leftarrow 0$;

\WHILE{ (exitFlag $== 0$)}

  \STATE  exitFlag $\leftarrow$ REPRECESS()

\ENDWHILE

\STATE Periodically call CLEAN().

\ENDFOR

\end{algorithmic}
\end{algorithm}

\subsubsection{Subroutines}



     Subroutine PROCESS  ensures that
all the coordinates of $\alpha_j$
satisfy the bound constraint conditions in CS-Ranker model (\ref{EQ_ite_CCCP_dual}).
    It initializes $\alpha_{i_0}$ with $i_0$ the index of the chosen  PSM  and
 updates the coordinates  $\alpha_j$  with bounds $A_j$ or $B_j$ changed  (Line 1-2).
Then it updates  gradient vector $g_j$, $j\in S$ (Line 3), where
$g$ is defined by
\begin{equation}\label{EQ_g_j}
    g_i \stackrel{\triangle}{=} \frac{\partial G(\alpha)}{\partial \alpha_i} = y_i - \sum_{j\in S}^{} \alpha_{j}^{}k(x_i,x_j).
\end{equation}



\bigskip

\begin{algorithm}
\caption{PROCESS}
\begin{algorithmic}[1]

\STATE    $\alpha_{i_0}^{}\leftarrow 0$  for new chosen PSM indexing $i_0$.
\STATE    For all $j\in S$ that bounds $A_j$ or $B_j$ has been changed,
 update $\alpha_j$:   $\alpha_j^{} \leftarrow 0$,

\STATE For all $j\in S$,  calculate $g_j$:
    $
    g_j \leftarrow y_j - \sum_{s\in S}^{} \alpha_{s} k(x_j,x_s). 
    $
\end{algorithmic}
\end{algorithm}


    Subroutine REPROCESS aims 
    to    find a better solution of  model (\ref{EQ_ite_CCCP_dual}).
It selects the instances with
the maximal gradient in active set $S$(Line\,1 -- Line\,12).
Once an instance is selected, it computes a stepsize (Line\,13 -- Line\,17) and
performs a direction search (Line\,18 -- Line\,19).

\bigskip


\begin{algorithm}
\caption{exitFlag $=$ REPROCESS()}
\begin{algorithmic}[1]
\STATE $i\leftarrow \textnormal{argmin}_{s\in S}^{} g_s $ with $\alpha_s > A_s^{}$;

\STATE $j\leftarrow \textnormal{argmax}_{s\in S}^{} g_s $ with $\alpha_s < B_s^{}$.

\IF{ $ \max(g_j,  - g_i) \leq  \tau$ }
\STATE   exitFlag $= 1$;
   \textbf{Return;}

\ELSE
\STATE  exitFlag $= 0$;
\ENDIF

\IF{  ($- g_i > \tau$, $g_j < \tau$ ) Or  ($- g_i > \tau$, $g_j > \tau$ and $-g_i > g_j  $) }
 \STATE  $g\leftarrow g_i$, $t \leftarrow i$;
\ELSE
\STATE  $g\leftarrow g_j$, $t \leftarrow j$;
\ENDIF

\IF{ $g<0$ }

\STATE $\lambda = \max(A_t-\alpha_t,\frac{g}{K_{tt}})$

\ELSE
\STATE $\lambda = \min(B_t-\alpha_t,\frac{g}{K_{tt}})$
\ENDIF

\STATE $ \alpha_t^{} \leftarrow \alpha_{t}^{} + \lambda $,
\STATE $g_s \leftarrow g_s -\lambda K_{is}^{}$, $\forall s\in S$.
\end{algorithmic}
\end{algorithm}


 Subroutine CLEAN     \red{removes}    PSMs that are not  effective to the discriminant function    from the active set $S$
 to minimize the requirement of memory and computation.
The  subroutine selects  non-support vectors as a set $V$
   (Line\,\ref{alg_clean_initial} -- Line\,\ref{alg_clean_initial_end}),
and then remove
\blue{$m$}  PSMs
among $V$ 
with largest gradients (Line\,\ref{alg_clean_remove_start} -- Line\,\ref{alg_clean_remove_end}).


\begin{algorithm}
\caption{CLEAN}

\textbf{parameter:}

 $m$: maximum number of removed non-support vectors;


\begin{algorithmic}[1]

\STATE\label{alg_clean_initial}   $V \leftarrow \emptyset$
\FOR{ $i$: $i\in S$, $\alpha_i=0$}
\STATE
         $V\leftarrow V \cup \{ i \}.$
\ENDFOR \label{alg_clean_initial_end}

\IF{ $|V| \leq m$} \label{alg_clean_remove_start}
\STATE remove $i$ from $S$,   $\forall \, i\in V$
\ELSE
\STATE select $m$ indices from $V$ with largest gradients $g_i$ and remove from $S$.
\ENDIF\label{alg_clean_remove_end}

\end{algorithmic}
\end{algorithm}


\subsubsection{Calculate PSM scores} 

After  discriminant function $\hat{f}$:
  $  \hat{f}(x) = \sum_{j\in S}^{} \alpha_j k(x_j,x),$
 where $k(\cdot)$ is the selected kernel function, is trained, we calculate
the scores of all PSMs on both training and test sets.
 The score of PSM $(x_i,y_i)$ is defined in \cite{Liang2015cranker}:
$$
    score(i) = \frac{2}{\pi}\arctan(\hat{f}(x_i)).
$$
The larger the score value is, the more likely a PSM is  correct.
The PSMs are ordered according to their scores, and a certain
number of PSMs are reported according to a pre-selected FDR.

\section{Results and discussion}

\subsection{Experimental Setup}

To evaluate the performance of OLCS-Ranker, we compare its performance against other algorithms on  eight  MS/MS datasets:
  universal proteomics standard set (Ups1), the \textit{S.\,cerevisiae} Gcn4 affinity-purified complex (Yeast), \textit{S. cerevisiae} transcription complexes using the Tal08 minichromosome (Tal08 and Tal08-large) and Human Peripheral Blood Mononuclear Cells  (PBMC datasets: Orbit-mips, Orbit-nomips, Velos-mips and Velos-nomips).
The MS/MS spectra were extracted from the mzXML file using the program MzXML2Search and all data was processed using the SEQUEST software. Refer to \cite{jian13} for the details of the sample preparation and LC/MS/MS analysis.
 For analysis of SEQUEST output with PeptideProphet,
we  used the Trans Proteomic Pipeline V.4.0.2 (TPP).
For analysis of SEQUEST output with Percolator, we  converted the SEQUEST outputs to a merged file in SQT format \cite{Mcdonald2004,SQT}. 
OLCS-Ranker
 was implemented with Matlab  R2015b and ran on  a PC with
 Intel Core E5-2640 CPU 2.40GHz  and 24Gb RAM.

Statistics of the SEQUEST search results of the  eight  datasets are listed in Table~\ref{Tab_detail_sequest},
where ``Full'', ``Half'' and ``None'' indicates
 the three types of tryptic peptides: full-digested, half-digested and none-digested peptides
 generated by Trypsin digestion of the protein samples.

\begin{table}[htb]
  \caption{\label{Tab_detail_sequest}  Statistics of datasets.
   }
   \begin{small}
  \begin{tabular}{l*{4}{c}c@{\hspace{0.3ex}}*{3}{c}}
  \toprule
        &  \multirow{2}*{Total}
        &   \multicolumn{3}{c}{Target}& &  \multicolumn{3}{c}{Decoy}  \\ \cline{3-5} \cline{7-9}
        &   & Full & Half & None      &   & Full & Half & None     \\
  \midrule
  Ups1          &17335   & 645 	   	&2013 &6316    &  &236  &2588  &5537   \\
  Yeast         &14892   & 1453	   	&1210 &4040    &  &106  &1465  &6618   \\
  Tal08         &18653   & 1081  	&2133 &6693    &  &164  &1923  &6659   \\
  Tal08-large   &69560   & 14893 	&6809 &20520   &  &419  &5877  &21042   \\
  Orbit-mips    &103679  & 26760  	&15647 &25927  &  &737  &8583  &26025   \\
  Orbit-nomips  &117751  & 28561  	&17490 &30344  &  &948  &10333  &30075   \\
  Velos-mips    &301879  & 110404  	&35915 &62446  &  &2520  &24682  &65912  \\
  Velos-nomips  &447350  & 134117  	&77052 &96380  &  &3414  &34985  &101402  \\
  \bottomrule
  \end{tabular}
  \end{small}
\end{table}

  Following CRanker \cite{Liang2015cranker}, a  PSM record is represented by a vector of nine attributes: xcorr, deltacn, sprank, ions, hit mass, enzN, enzC, numProt, deltacnR.
   PSMs \red{have} randomly divided into a training set and  a test set according to the ratio 2:1.
       The numbers of  PSMs identified on the training set and on the test set are calculated.
Weight 1.0 was assigned for xcorr and deltacn, and 0.5 for all others.
We  used   the Gaussian kernel $k(x_i,x_j) = \exp{(\frac{\|x_i-x_j\|^2}{2\sigma^2})} $
  for OLCS-Ranker   with kernel parameter $\sigma=1$.

We choose the values of parameters of OLCS-Ranker
by 3-fold cross-validation in terms of
  the number of identified PSMs and test/total ratio.

\subsection{Comparison with benchmark methods}

We compared OLCS-Ranker, PeptideProphet  and Percolator on the seven datasets, and
Table\,\ref{Tab_PSM_ranker_compare} shows the numbers of validated PSMs  at FDR $\approx 0.05$.
The performance of a validation approach is better if it validates more target PSMs compared to another approach with the same FDR.
As we can see, OLCS-Ranker identifies more PSMs than PeptideProphet and Percolator over
 all the  eight  datasets.
Particularly, 5\% more PSMs were identified by OLCS-Ranker on Tal08 and Tal08-large, and the improvement on the other datasets were about 2\% or more.

\begin{table}[!h]
  \caption{\label{Tab_PSM_ranker_compare} Target PSMs output by PeptideProphet, Percolator, and OLCS-Ranker.
     \#TP: number of true positive PSMs.
     \#FP: number of flase positive PSMs.
     \#(total targets): number of the total targets in the PSM dataset (refer to Table\,\ref{Tab_detail_sequest}
     for detailed quantities.)
   }
  \begin{small}
  \begin{tabular}{ll*{1}c*{5}c}
  \toprule
  Data set  & Method   &  Total & \#TP   &  \#FP    & \#TP/\#(total targets) \\
  \midrule
  \multirow{3}*{ups1}    & PepProphet &582     &566     &16      & 6.3\%       \\
                         & Percolator & 450    &438     &12   	 & 4.9\%        \\
                         & OLCS-Ranker  & 597    &582     &15    & 6.5\%       \\      \midrule
  \multirow{3}*{yeast}   & PepProphet &1481    &1443    &38      & 21.5\%      \\
                         & Percolator & 1429   &1394    &35   	 & 20.8\%      \\
                         & OLCS-Ranker    & 1516  &1479    &37   & 22.1\%	\\    \midrule
  \multirow{3}*{tal08}   & PepProphet &982   &957   &25  	 & 9.7\%       \\
                         & Percolator &978   &953   &25    	 & 9.6\%       \\
                         & OLCS-Ranker    &1145   &1118   &27 	 & 11.3\%      \\     \midrule
  \multirow{3}*{tal08-large}&PepProphet &16025   &15638   &387   & 37.0\%      \\
                         & Percolator &14725   &14371   &354     & 34.0\%      \\
                         & OLCS-Ranker    &16792   &16373   &419 & 38.8\%      \\  \midrule
\multirow{3}*{Orbit-mips}  & PepProphet &34035   &33233   &802 	 & 48.6\%      \\
                           & Percolator &34118 &33270 &848   	 & 48.7\%      \\
                           &OLCS-Ranker    &35095   &34257  &838 & 50.1\%      \\ \midrule
\multirow{3}*{Orbit-nomips}&PepProphet  &36542   &35673  &869 	 & 46.7\%      \\
                           & Percolator &36962   &36096   & 866  & 47.2\%      \\
                           & OLCS-Ranker   &37387   &36454  &933 & 47.7\%      \\ \midrule
\multirow{3}*{Velos-mips}  & PepProphet &123908  &120961 &2947	 & 57.9\%      \\
                           & Percolator &125701  &122568 &3133   & 58.7\%      \\
                           & OLCS-Ranker  &126015  &122866 &3149 & 58.9\%      \\ \midrule
\multirow{3}*{Velos-nomips}& PepProphet &180182  &175789 &4393	 & 57.2\%      \\
                           & Percolator &178280  &173917 &4363   & 56.5\%	\\
                           & OLCS-Ranker  &183573  &178985 &4588 & 58.2\%      \\
  \bottomrule
  \end{tabular}

  \smallskip

\quad {\footnotesize PepProphet: PeptideProphet.  \hfill{}\mbox{}}
  \end{small}
\end{table}

\begin{figure}[!t]
  \includegraphics[width=1.01\columnwidth]{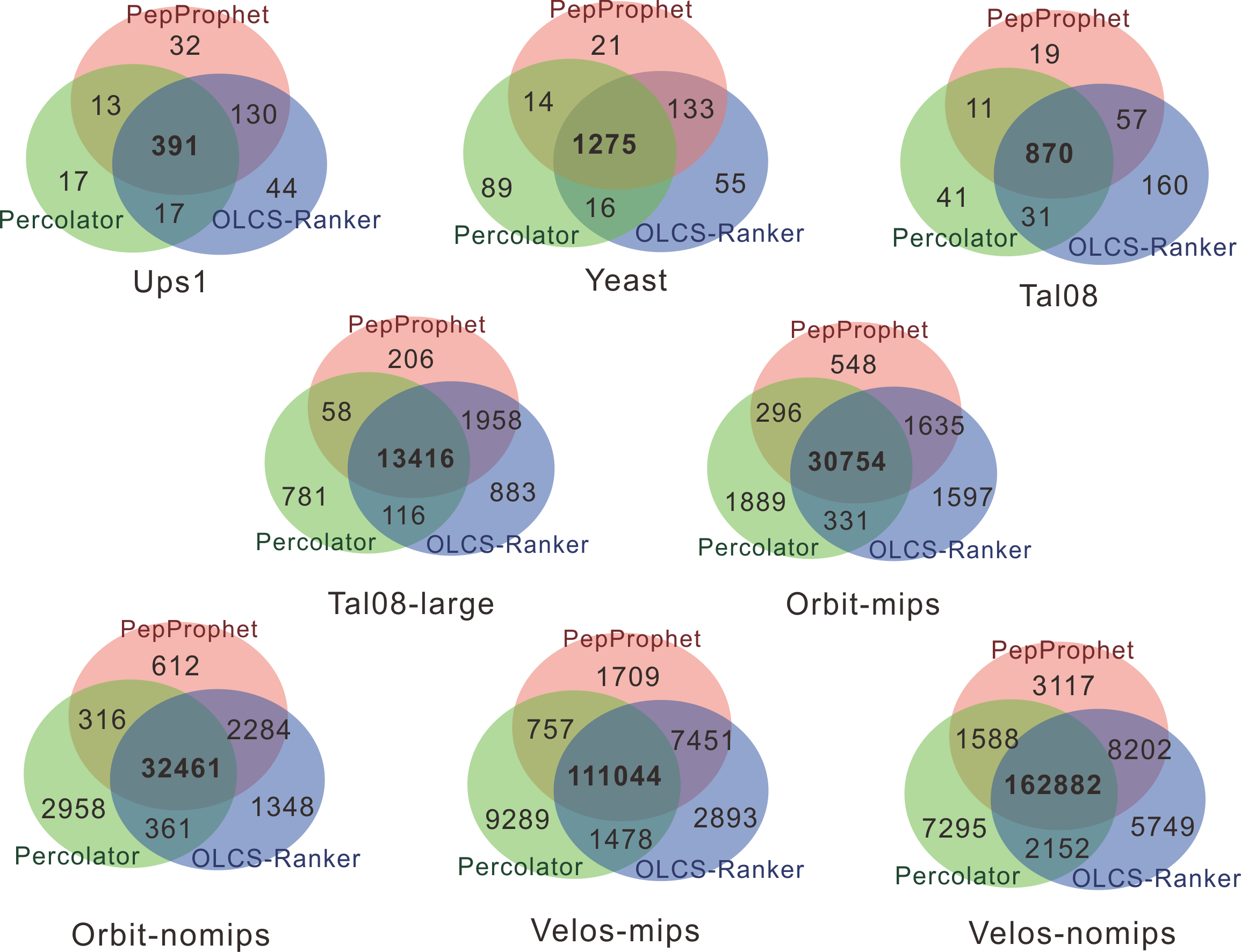}
    \caption{\label{Fig_overlap}Overlap of identified target PSMs by PeptideProphet, Percolator and OLCS-Ranker.
         PepProphet: PeptideProphet.  }
\end{figure}

We also compared the overlapping of target PSMs identified by OLCS-Ranker, PeptideProphet and Percolator,
as a PSM reported by multiple methods is  more likely to be correct.
As we can \red{see} in Figure\,\ref{Fig_overlap}, the majority of validated PSMs by the three approaches overlaps.
For example,
on Tal08, the three algorithms have 870 PSMs in common,
covers more than 77.8\% of the total target PSMs identified by each of the \red{algorithms}.
This   ratio of common PSMs is 86.2\%  and 81.9\% on Yeast and Tal08-large, respectively,
    and more than 90\% on the  four PBMC datasets.
Moreover,  on each dataset,  OLCS-Ranker identified a certain part of  PSMs that identified by PeptideProphet but not by Percolator,
or identified by Percolator but not by PeptideProphet.
The above overlap results \red{indicate} that the identified PSMs output by OLCS-Ranker are reasonable.

\textbf{Hard datasets and normal datasets}

  As shown in Table\,\ref{Tab_PSM_ranker_compare},  PeptideProphet, Percolator and OLCS-Ranker reported similar ratios of the number of identified target PSMs to the total target PSMs over eight datasets.
 Note that the ratios on two datasets Ups1 and Tal08 are relatively lower than the other six datasets.
Particularly, the ratios on Ups1 and Tal08 are  4.9\%$\sim$6.5\%, 9.6\%$\sim$11.3\%  respectively while the ratios are
more than 20\% on the other six datasets.
As a large proportion of incorrect PSMs in a dataset usually reduces the accuracy of PSM identification, we named
two datasets Ups1 and Tal08 ``hard datasets''.
In contrast, the other six datasets  have much lower   ratios of incorrect matches than those of Ups1 and Tal08,
 and we named them ``normal datasets''.

\subsection{Model evaluation}

We use a separate test dataset to examine whether the OLCS-Ranker model overfits the training datasets.
The ratio of identified PSMs
to the total PSMs (test/total ratio) is calculated.
As the test datasets are randomly chosen from the whole datasets according to the ratio of 1:3,
a value of 33.3\% is an expected test/total ratio.
The test/total ratios of all datasets  under FDR$\approx$0.05
are listed in  the last column of Table\,\ref{Tab_fdr_test}.

\begin{table}[!t]
   \caption{\label{Tab_fdr_test}   FDR of  OLCS-Ranker on test set.
      $\frac{test}{total}$: test/toal ratio, the ratio of PSMs identified on the test set to that on the  total dataset.
      FDR: false discovery rate.
   }
   \begin{center}
  \begin{tabular}{l*{4}{l}}
  \toprule
  &  \#TP  & \#FP &   FDR     & $\frac{test}{total} $  \\
  \midrule

  {Ups1}        &180       &5       &5.41\%  & 30.99\% \\
  {Yeast}       &491          &12         &4.77\%  & 33.18\% \\
   {Tal08}       & 348      &8          &4.49\%   & 31.10\% \\
   {Tal08-large}      & 5546      &139          &4.98\%   & 33.26\% \\
 {Orbit-mips}    &11384   &281     &4.99\%   & 33.24\% \\
 {Orbit-nomips}  &12048   &308    &4.99\%   & 33.05\% \\
 {Velos-mips}    &40905   &1048   &5.00\%   & 33.29\% \\
 {Velos-nomips}  &59777 &1532 &5.00\%    & 33.39\% \\
  \bottomrule
  \end{tabular}
  \end{center}
\end{table}

\begin{figure*}[htb]
  \includegraphics[width= \textwidth]{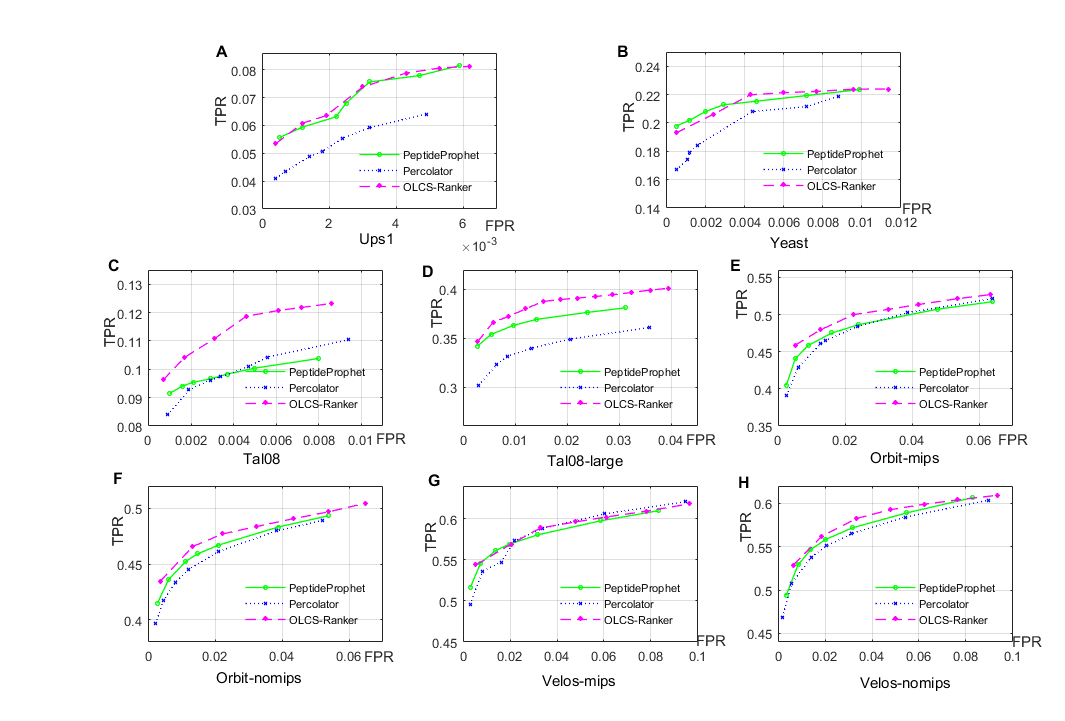}
  \caption{\label{Fig_roc} ROC curves.}
\end{figure*}

On all the six ``normal datasets'', the test/total ratios are extremely close to the ideal ratio,
indicating that no overfitting problem  occurs  on OLCS-Ranker classifiers.
On the remaining two ``hard datasets'', Ups1 and Tal08, the test/total ratio is  near 31\%,
which is 2.3\%   lower than
the ideal ratio.
 The relatively low test/total ratios on ``hard datasets'' is mainly induced  by extremely unbalanced PSM datasets, in which there are few correct target PSMs.

We have also evaluated the performance of OLCS-Ranker, PeptideProphet and Percolator by using
receiver operating characteristic (ROC).
As shown in Figure\,\ref{Fig_roc},
OLCS-Ranker reaches highest TPRs among the three methods
at most values of FPRs on the eight datasets, except the
FPR at  0.003 on Ups1, the FPR in the range from 0.001 to 0.003
on Yeast, and the FPR in the range from 0.05 to 0.1 on Velos-mips.
Particularly, CS-CRanker reaches significantly higher TPR levels than that
of PeptideProphet and Percolator on Tal08  and Tal08-large  dataset.

\subsection{The  algorithm  efficiency}

To evaluate the efficiency of OLCS-Ranker, we compare its algorithmic performance with those of
  Batch-CS-Ranker and C-Ranker.
Batch-CS-Ranker  solves model (\ref{EQ_cranker_1var})
by using Algorithm\,\ref{Alg_Batch_CS_Ranker}  which transforms  subproblem (\ref{EQ_ite_CCCP}) to its dual programming and
then solved it with Matlab built-in solver while OLCS-Ranker solves  model (\ref{EQ_cranker_1var}) by using Algorithm\,\ref{alg_OLCS_Ranker}, which is our designed online algorithm.
As the Batch-CS-Ranker algorithm and C-Ranker require the whole training data for construction of kernel matrix, it
 is difficult to implement the two algorithms  on large-scale datasets. 
 Instead, we divided the whole training dataset into five subsets by randomly selecting 16000 PSMs for each subset.
The final score of a PSM is the average of the scores on the five subsets.

\begin{table}[]
\centering
\caption{Comparing OLCS-Ranker with CRanker and Batch-CS-Ranker algorithm}
\label{table_online_batch_runtime}
\begin{tabular}{lllll}
\toprule
Dataset      & Method           & \#Total PSMs   & $\frac{test}{total} $   &  time (s)  \\
\midrule
\multirow{2}{*}{Ups1}
			     &C-Ranker & 614.3    & 27.24\% & 1507.0   \\
                             & Batch-CS-Ranker   & 597.7         & 31.73\% & 255.4   \\
                             & OLCS-Ranker        & 590.6        & 30.77\% & 19.3   \\\hline

\multirow{2}{*}{Yeast}
 			     &C-Ranker 		 & 1462.3   & 33.90\% & 667.8     \\
                             & Batch-CS-Ranker  & 1489.3        & 31.81\% & 642.5  \\
                              & OLCS-Ranker        & 1507.9      & 31.89\% & 16.9   \\\hline
\multirow{2}{*}{Tal08}
 			     &C-Ranker & 1118.7   & 30.61\% & 1579.6   \\
                             & Batch-CS-Ranker  & 1116.7     & 30.85\% & 345.3   \\
                             & OLCS-Ranker        & 1146.5        & 30.28\% & 26.0    \\\hline
\multirow{2}{*}{Tal08-large}
 				&C-Ranker & 16659.3  & 33.39\% & 10090.1   \\
                               & Batch-CS-Ranker  & 16366.0     & 33.28\% & 8088.3  \\
				 &OLCS-Ranker & 16725.7  & 33.07\% & 116.7  \\\hline
\multirow{2}{*}{Orbit-mips}
				 &C-Ranker & 34720.3  & 32.99\% & 10207.5   \\
                                & Batch-CS-Ranker  & 34557.7     & 33.29\% & 18264.0  \\
				 &OLCS-Ranker & 35222.3  & 33.07\% & 146.2  \\\hline
\multirow{2}{*}{Orbit-nomips}
				 &C-Ranker & 37147.3  & 33.25\% & 9630.1    \\
                              & Batch-CS-Ranker  & 36738.0     & 33.30\% & 22428.1  \\
 				&OLCS-Ranker & 37321.7  & 33.21\% & 155.8  \\\hline
\multirow{2}{*}{Velos-mips}
 				&C-Ranker & 125435.0 & 33.40\% & 9052.9    \\
                                & Batch-CS-Ranker  & 124233.0    & 33.34\% & 21107.0  \\
				 &OLCS-Ranker & 125328.7 & 33.33\% & 495.5  \\\hline
\multirow{2}{*}{Velos-nomips}
			       &C-Ranker & 182665.7 & 33.18\% & 11478.5   \\
                               & Batch-CS-Ranker  & 179811.7    & 33.31\% & 23849.7  \\
				 &OLCS-Ranker & 182276.3 & 33.32\% & 754.3 \\ 	
             \bottomrule
\end{tabular}

\end{table}

Table\,\ref{table_online_batch_runtime} summaries performance comparison among OLCS-Ranker, C-Ranker and Batch-CS-Ranker in terms of \red{the} total number of identified PSMs, total/test ratio, elapsed time.
 We 
 measured the execution time of each algorithm on eight datasets.
As we can see in Table\,\ref{table_online_batch_runtime}, OLCS-Ranker is about $15\sim 85$ times  faster than CRanker and
 $30\sim 140$ times  faster than Batch-CS-Ranker algorithm on large-scale datasets.

\subsection{The  algorithm  stability}

\begin{figure}[!t]
  \includegraphics[width=\columnwidth]{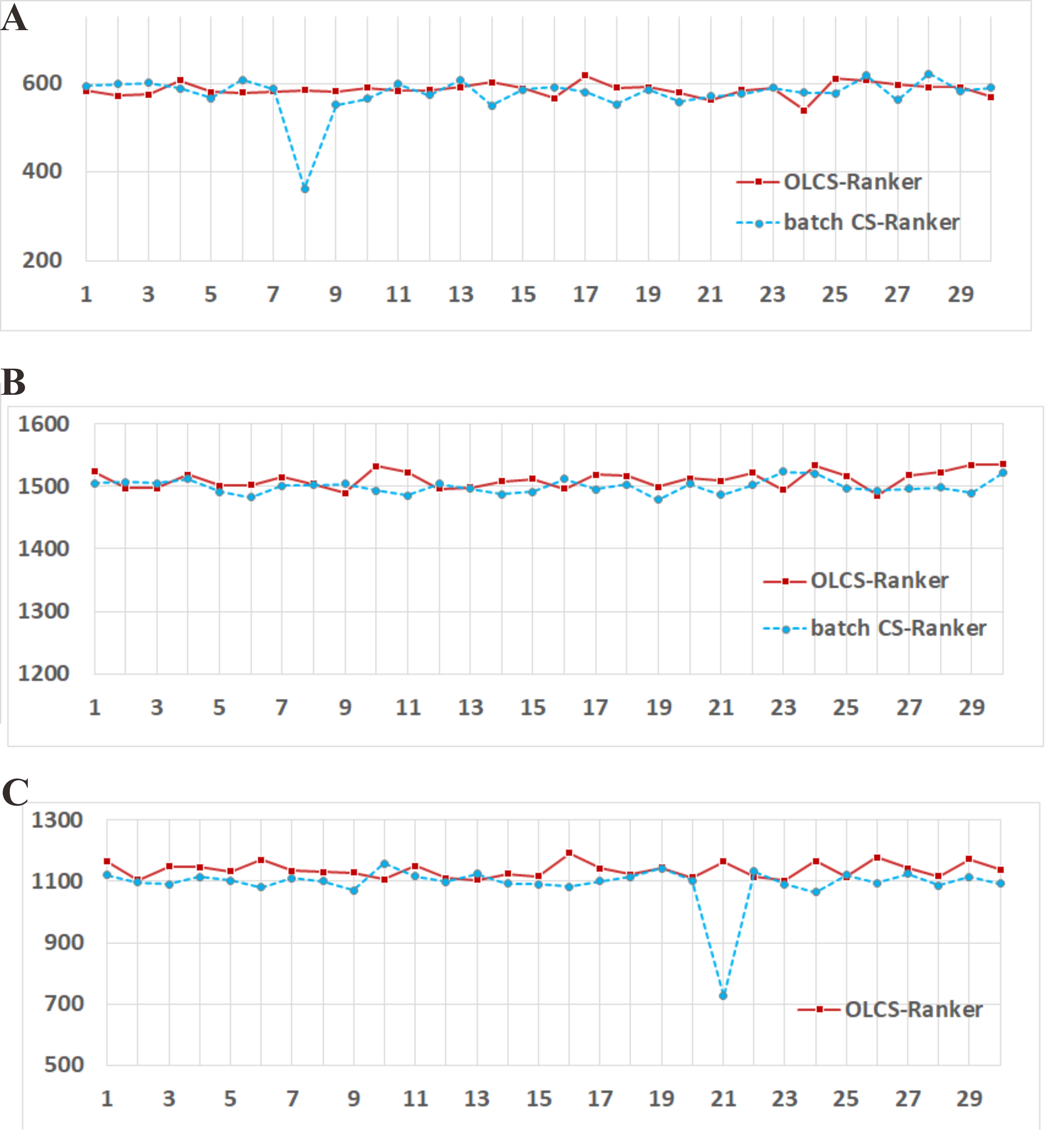}
   \caption{Numbers of PSMs   identified by OLCS-CRanker and Batch-CS-Ranker algorithm in 30 times}
        \label{Fig_stability}
\end{figure}

As the training data are randomly selected from the training dataset, the output of a model-based algorithm may slightly vary due to different training PSMs.
Ideally, a good algorithm should report very similar target PSMs even with \red{a} different training set.
 We have run OLCS-Ranker and Batch-CS-Ranker for 30 times,
 each with an independent training set.
 Due to the excessive
 computation of Batch-CS-Ranker on large-sized datasets, we compared the
 stability   on three small datasets.  
  The numbers of identified PSMs at the 30 runs on Ups1, Yeast and Tal08 
   are depicted in Figure\,\ref{Fig_stability} (A), (B) and (C), resp.
Note that Batch-CS-Ranker reported a relatively small number of identified PSMs  at the 8-th trial  on Ups1
(Figure\,\ref{Fig_stability} (A)) and the 21-th trial on Tal08 (Figure\,\ref{Fig_stability} (C)), which indicates the  optimization solver used in Batch-CS-Ranker
 is trapped in  bad local minima.
  In contrast, OLCS-Ranker, with online iteration technique,
can escape from the bad local minima.

\section{Conclusion}
We have presented a cost-effective post-database search approach
for peptide identification using \red{an} SVM-based learning model, which
introduces different penalties for learning errors
on decoy an target PSMs.
An efficient online learning algorithm, OLCS-Ranker, is designed
for tackling the challenge of identification on hard datasets and
large-scale datasets.
Experimental studies show that OLCS-Ranker based on the cost-effective learning model
has improved the learning algorithmic performance and increased identified PSMs,
compared with
other kernel-based batch algorithms,  CRanker and Batch-CS-Ranker.

%

\section*{Funding}

This study was supported the National Natural Science Foundation of China under Grant No.\,61503412
and the Fundamental Research Funds for the Central Universities under Grant No.\,16CX02048A.

\section*{Notes}
The authors declare no competing financial interest.

%
%
%
%
%
%
%
%

%
\bibliographystyle{plain}
%

\bibliography{olcs-ranker}

\end{document}